# Detection of Plant Leaf Disease Directly in the JPEG Compressed Domain using Transfer Learning Technique

Atul Sharma[1], Bulla Rajesh[2*] and Mohammed Javed[3]

Computer Vision and Biometrics Laboratory (CVBL),
Department of Information Technology
Indian Institute of Information Technology Allahabad, Prayagraj, U.P, India

[1]`ism2016003@iiita.ac.in`, [2]`rajesh091106@gmail.com`,
[3]`javed@iiita.ac.in`

**Abstract.** Plant leaf diseases pose a significant danger to food security and they cause depletion in quality and volume of production. Therefore accurate and timely detection of leaf disease is very important to check the loss of the crops and meet the growing food demand of the people. Conventional techniques depend on lab investigation and human skills which are generally costly and inaccessible. Recently, Deep Neural Networks have been exceptionally fruitful in image classification. In this research paper, plant leaf disease detection employing transfer learning is explored in the JPEG compressed domain. Here, the JPEG compressed stream consisting of DCT coefficients is, directly fed into the Neural Network to improve the efficiency of classification. The experimental results on JPEG compressed leaf dataset demonstrate the efficacy of the proposed model.

**Keywords:** Deep Neural Networks, Transfer learning, Classification, DCT coefficients, JPEG compressed domain.

## 1   Introduction

The need for food is growing day by day as the human population grows. According to the United Nations, the Human population may reach 9.7 billion in 2050 [1]. Considering this growth rate, it is very important to minimize food loss especially in overpopulated countries like India. In a country where farming remains an important backbone for economic growth, it is very important to protect crops from being damaged due to leaf diseases. For sustainable food production, therefore, plant disease identification and management are important. The majority of food crop losses are due to different diseases in the plant, and about 80% to 90% of plant diseases occur on the plant leaf [2]. Therefore in this research work, our focus is on detecting the disease on the leaf, rather than the whole plant.

   Disease identification in plants assumes a significant part in agribusiness as farmers need to choose whether the yield they are reaping is adequate. It is of most extreme importance to pay attention to these as they can prompt difficult issues in plants because of which individual item quality, amount, or profitability is influenced. The crop should be monitored for sickness from the first absolute step of its life cycle to the point of collection. The technique for screening plants for diseases was initially the usual unassisted eye vision which is an awkward process that requires specialists to foresee [10]. Automatic characterization of plant illnesses is significant during the beginning phases; this would help in identifying the side effects of the illness when they show up on plant leaves. Therefore this research aims to develop an automatic leaf disease classifier using computer vision technique.

*Corresponding author

In recent years, Deep learning techniques have led to significant classification results. In this research work, a transfer learning-based technique is explored for leaf disease detection. On the other hand, we know that any digital image needs to be stored in the compressed form, and also decompressed later for further processing. Therefore in the literature, there have been techniques that directly feed the compressed image data into the neural network and achieve the same classification task very efficiently [3][15][16]. Document analysis is one of the areas where the compressed domain has been widely explored [17][18]. This motivates us to explore the problem of leaf disease detection directly in the compressed domain. Due to the popularity and efficiency of the JPEG compression algorithm, the digital images are generally stored in the JPEG compressed format. In the present research work, JPEG compressed leaf image dataset is used for experimentation.

The rest of the paper is arranged as follows. Section 2 covers the literature relevant to leaf disease detection, section 3 discusses the proposed model and deep learning architecture, section 4 reports about the dataset, experiments, and results, and section 5 briefly summarizes the research work.

## 2  Related Literature

Most of the diseases in plants start from the leaves, and they are the part of the plant where indications of most of the diseases appear first. (D.A. Bashish et al.,)[4] used K-means which uses squared Euclidean distances for segmenting the images into clusters. The color co-occurrence approach, which combines picture color and texture, is used to extract features. The classification is carried out with the help of neural networks based on the backpropagation technique. The accuracy was 93% and a total of 5 diseases were classified. (Al-Hiaryet et al.,)[5] added an additional step of masking the green pixels after applying the K-means clustering algorithm. The threshold value is calculated based on Otsu's method [13]. The red, green, and blue integrants of the pixel having a green component value less than the threshold value are set to zero. Further, the constituents which are on the border of the diseased cluster and those having zero red, green, and blue values are removed. The color co-occurrence approach is used to extract features and the classification is done using neural networks. The 5 diseases were classified with an accuracy of 94%. (Revathi and Hemalatha)[6] extracted three features which are color variance, shape variance, and texture variance using color histogram, Sobel and canny filters, and Gober filter respectively. The feature selection is done by using Particle Swarm Optimization. It classifies the 270 samples into 6 classes with an accuracy of 95 %. ( Deshpande et al.,)[7] detected Bacterial Blight disease on the pomegranate leaves. The diseased area is extracted by image segmentation through K-means clustering and if the extracted area is bounded by a yellow color, then the leaf is classified as infected by bacterial blight. (Gavhale et al.,)[8] classified citrus fruit leaf diseases canker and anthracnose. The author first does an image enhancement by scaling the Discrete Cosine Transform coefficients and converting the image into YCbCr color space, and then image segmentation is done by using K-means, and texture features are extracted by GLCM. Final classification was done by the SVM and the accuracy of 95% on 300 sample images was reported. (G. Saradhambal et.al.,)[9] converted the RGB image into Lab color space before the segmentation. The image segmentation is done by using K-means and Otsu's algorithm. The shape (perimeter, area, etc.) and textural (contrast, energy, etc.) features were extracted to classify the image.

In the traditional approaches like SVM, ANN, etc., the feature extraction is done through GLCM, histograms, CCM, etc., while in deep learning the models do the automatic feature extraction due to which deep learning has outshined the traditional approaches(Andreas et al.,)[10]. This research work is motivated to use a deep transfer learning approach for plant leaf disease detection.

## 3 Proposed Methodology

This section describes JPEG compression's background details and the background details of JPEG compression and also discusses the proposed deep neural architecture for plant leaf disease detection in the JPEG compressed domain.

### 3.1 Background of DCT Compression

This section provides a brief explanation of the JPEG algorithm and compressed input representation. Each image in the RGB color space is converted to YCbCr format. This conversion aids in distinguishing between Luminance and Chrominance.

After that, each block of 8 × 8 pixels of the image, is converted into a frequency domain by applying discrete cosine transformation (DCT) to each of the channels (Y, Cb, Cr). The DCT coefficient $D(i,j)$ is calculated by equation 1 below.

$$D(i,j) = \frac{C_i C_j}{4} \sum_{x=1}^{7} \sum_{y=1}^{7} p(x,y) \cos\cos\left[\frac{(2x+1)i\pi}{16}\right] \cos\cos\left[\frac{(2y+1)i\pi}{16}\right] \quad (1)$$

where,

- $C_i$ and $C_j$ are the horizontal and vertical spatial frequency having values between 0 and 7 respectively
- $p(x,y)$ is the pixel value at coordinate $(x,y)$

The DC values obtained by applying DCT are sorted in increasing order of the frequencies. It is widely known that low-frequency components have a higher sensitivity in human eyes than high-frequency components. Therefore we remove many of the high-frequency details through the process of quantization (equation 2) where we divide each element of the transformed matrix D by each element of the quantization matrix and round to the nearest integer value.

$$C_{i,j} = \text{round}\left(\frac{D_{i,j}}{Q_{i,j}}\right) \qquad (2)$$

Where,
- $D_{i,j}$ and $Q_{i,j}$ are the corresponding DCT and quantization values respectively
- $C_{i,j}$ is the quantized value

The matrix obtained after the quantization will contain many zeroes. Therefore the matrix is re-ordered from the top left corner by moving in a zigzag manner. At the end of this step, we will get the number of zeros at the end.

The DC coefficients are coded using the DPCM because they are large and varied but near to previous values. If $DC_i$ is the DC coefficient then the corresponding DPCM value $d_i$ is given by equation 3 below.

$$d_i = DC_{i+1} - DC_i \qquad (3)$$
$$d_0 = DC_0$$

Then we apply run-length encoding on the vector obtained after zigzag scanning. We store each of the non-zero values along with the number of zeros before it. This converts the image into binary form.

The decompression is done by doing the inverse of each of the steps described above in the opposite order. So the decompression starts with the entropy decoding and continues to change the run lengths to a sequence of zeros and coefficients. Coefficients are de-quantized to get the decompressed dct image. The flow diagram of how an input representation to the proposed model is generated is shown in Fig. 1. below.

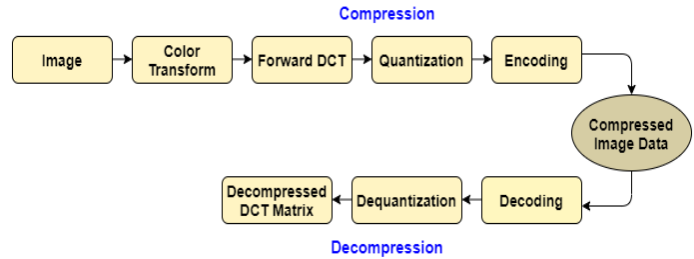

**Fig. 1.** Flow Diagram of the JPEG Compression and Decompression steps

### 3.2   Transfer Learning

Standard deep learning and machine learning algorithms presume that training and validation data must be in the same domain and have the same feature space distribution, so if there is any change in the feature space of the data the models have to be reconstructed and training need to be done again from the scratch. In such cases of change in the domain of the data, if the transfer of knowledge (features learned from a particular data distribution) can be done from one domain to another, the training time may thus be reduced, and the problem of data scarcity can be solved.[14]. This transfer of knowledge is known as transfer learning where the "Convolutional Neural Network trained for a particular task is used again as the beginning for the model on the other task" [11]. The difference between transfer learning and traditional learning is depicted in Fig. 2. below.

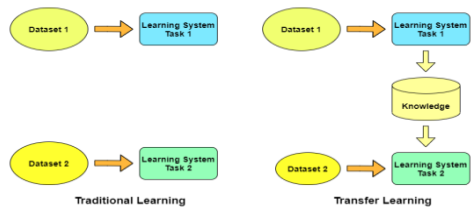

**Fig. 2.** Traditional Learning v/s Transfer Learning

### 3.3  Proposed Approach

The image is compressed using DCT, Quantization, and various encoding techniques into a binary format. The binary image is changed into decompressed DCT format (Fig. 3. below shows the RGB images and the corresponding converted decompressed DCT images in JPG format).

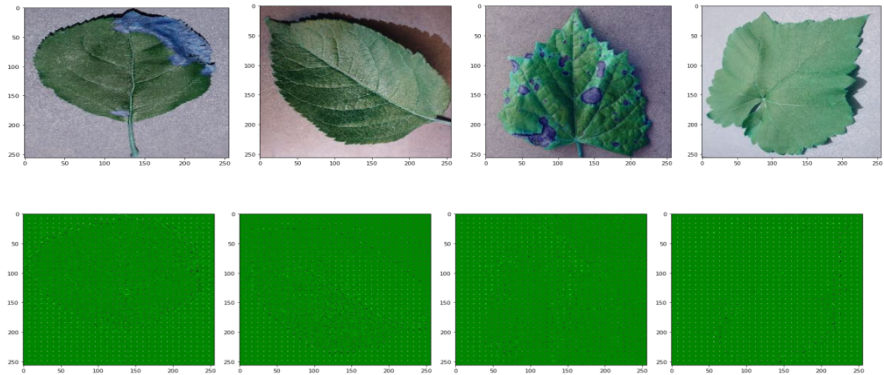

**Fig. 3.** The sample uncompressed disease plant leaves (top) and corresponding images in the DCT compressed domain (bottom)

The Convolution Neural Network is then trained on both the decompressed images and normal RGB images using the concept of Transfer Learning. The following Fig 4. below shows how the transfer learning concept is applied to solve our problem.

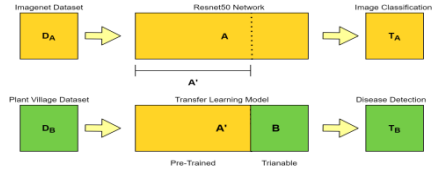

**Fig. 4.** A pictorial illustration of the Transfer Learning approach.

Our model is built on the ResNet50 architecture. The last layer of the ResNet50 is fine-tuned and on top of it, we have added a Global average pooling and two dense layers with 512 neurons each. The mean of the output of each of the previous layer feature maps is computed by the average pooling layer and the resulting vector is fed into the last softmax layer for the classification. The suggested approach is shown in Fig. 5.

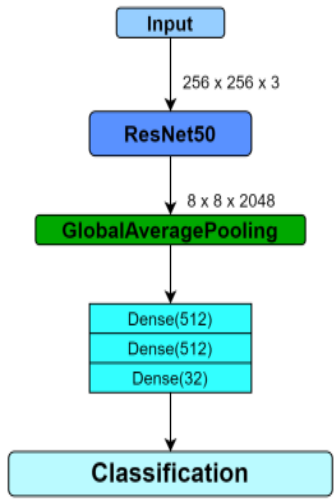

**Fig. 5.** The flow diagram of the proposed model for plant leaf disease detection in the JPEG compressed domain.

3.4  **Network Architecture**

We have used ResNet50 [12]. It is the 50 layer residual network through which we can train extremely deep neural networks. In a traditional neural network, the next layer input is the output of the previous immediate layer, but in the ResNet50, the output of each layer is provided as input to the next layer and the to the layer 2-3 hops apart. The layers in the standard neural networks learn the true output *f(x)* but layers in the residual network learn the residual *R (x) = f (x) - x*. This way, over-fitting is avoided, and also the problem of vanishing gradients is prevented, which eventually helps in the training of deep neural networks.

The network accepts as input an image with height and width a multiple of 32 and initially, the convolution operations and max-pooling operations are done by using the 7 x 7 and 3 x 3 as kernel sizes respectively. After that, the network is divided into 4 stages, each of which comprises three residual blocks and three layers in each block. The kernel size in all three layers of each residual block of Stage1 is 64, 64, and 128 respectively. As we go through the stages, the input size is cut in half and the channel width is doubled. A summary of the architecture is shown in Fig 6. below.

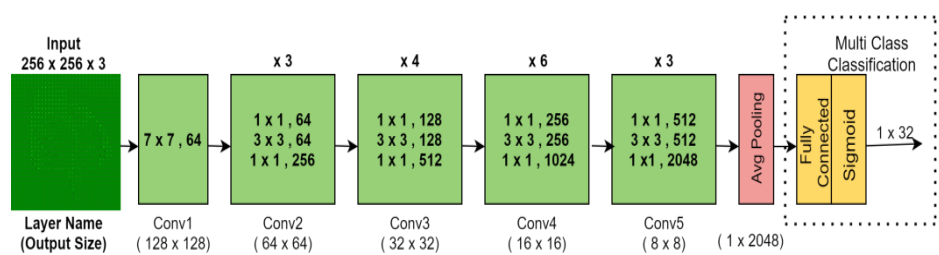

**Fig. 6.** The proposed deep architecture for plant leaves disease detection in JPEG compressed domain.

The last layers (fully connected and classification layer) of ResNet50 were removed; we have added a Global average pooling layer and two dense layers with 512 neurons each. The final classification is done by the last fully connected layer having 32 neurons and activation function as softmax.

## 4    Experimental Results

### 4.1    Dataset

The dataset utilized in the experiment includes 34,641 images of plant leaves. The 9 species of crops included are Apple, Corn, Grape, Cherry Peach, Pepper, Tomato, Potato, Strawberry. There are a total of 23 types of diseases out of which 16 are caused by fungi, 3 are caused by bacteria, 2 are caused by viruses, and 2 are caused by mold. The label of each class is a crop-disease pair. Each of the 32 classes along with their names is shown in Fig 7. below.

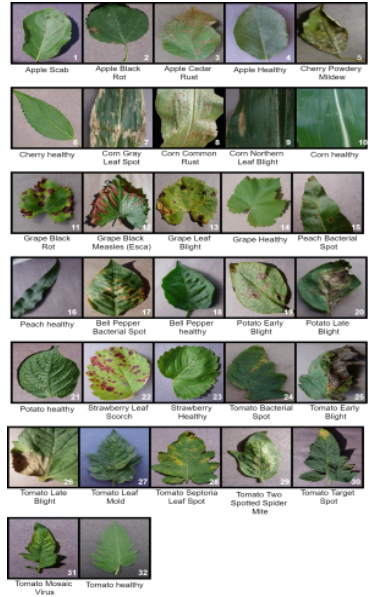

**Fig. 7.** The visualization of sample disease images for every crop-disease pair from the dataset.

### 4.2 Results

The model was trained using both normal RGB and decompressed DCT datasets. The accuracy and loss plots of both types of datasets are shown in Fig 8 and Fig 9 below.

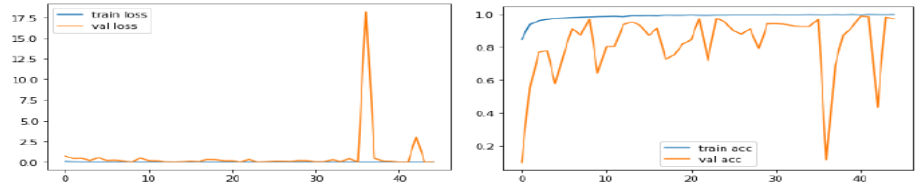

**Fig. 8.** A plot of loss and accuracy v/s epochs for plant leaves in the uncompressed domain.

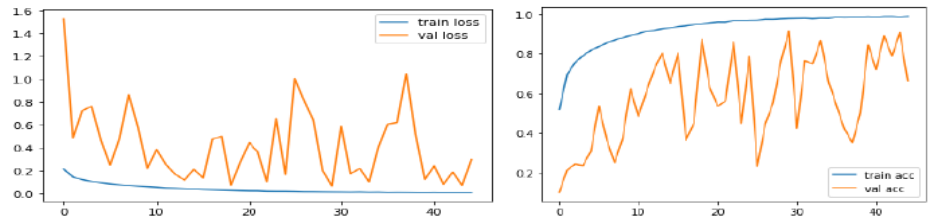

**Fig. 9.** A plot of loss and accuracy v/s epochs for plant leaves in the compressed domain.

The model's testing accuracy and Top3 Accuracy (accuracy when the true class matches any of the three most likely classes predicted) with various input types is shown in Table 1 below. The testing accuracy with Original RGB images is slightly more than Decompressed DCT images but the top3 accuracy is almost the same for both types of inputs. To get a clearer view of the miss-classification we have calculated the Precision(equation 4), Recall(equation 5), and F1-score(equation 6). The value of all of these metrics scores are shown in Table 2. below.

$$Precision = \frac{TP}{TP+FP} \quad (4)$$

$$Recall = \frac{TP}{TP+FN} \quad (5)$$

$$F1 = 2 * \frac{Precision \times Recall}{Precision+Recall} \quad (6)$$

**Table 1.** Experimental results of the suggested model with various kinds of inputs

| Model Type | Input Type | TestingAccuracy(%) | Accuracy (Top3) |
|---|---|---|---|
| **Transfer Learning** | Decompressed DCT Images | 92.63 | 98.94 |
| | Original Images (Pixel) | 98.49 | 99.48 |
| **Deep Learning** | Decompressed DCT Images | 85.58 | 94.55 |
| | Original Images (Pixel) | 92.06 | 95.46 |

**Table 2.** Experimental results of the suggested model with various kinds of inputs

| Model Type | Input Type | Precision | Recall | F1-Score | Training time(in min) |
|---|---|---|---|---|---|
| **Transfer Learning** | Decompressed DCT Images | 0.92 | 0.92 | 0.92 | 525 |
| | Original Images (Pixel) | 0.98 | 0.98 | 0.98 | 681 |
| **Deep Learning** | Decompressed DCT Images | 0.82 | 0.83 | 0.82 | 765 |
| | Original Images (Pixel) | 0.83 | 0.95 | 0.89 | 1020 |

The time required to train the model for the classification of Normal RGB images is longer than the time required for Decompressed DCT images because, during compression, the components of the image to which our eyes are not sensitive are removed, resulting in a smaller image size, which requires fewer computations and thus less time. The classification report for each of the 32 classes is shown in Table 3. below.

**Table 3.** Class-wise analysis of all the plant leaf diseases.

| Class Name | Precision | Recall | F1-Score |
|---|---|---|---|
| Scab(Apple) | 1.00 | 0.77 | 0.87 |
| Black Rot(Apple) | 1.00 | 1.00 | 1.00 |
| Rust(Apple) | 0.92 | 1.00 | 0.96 |
| Healthy(Apple) | 0.91 | 0.93 | 0.92 |
| Powdery Mildew(Cherry) | 0.96 | 0.98 | 0.97 |
| Healthy(Cherry) | 0.98 | 0.96 | 0.97 |
| Spot(Corn) | 1.00 | 1.00 | 1.00 |
| Rust(Corn) | 0.93 | 0.93 | 0.93 |
| Leaf Blight(Corn) | 1.00 | 0.98 | 0.99 |
| Healthy(Corn) | 0.67 | 1.00 | 0.80 |
| Blight(Grape) | 1.00 | 0.92 | 0.96 |
| Measles(Grape) | 0.93 | 0.92 | 0.92 |
| Black Rot(Grape) | 0.91 | 0.80 | 0.85 |
| Healthy(Grape) | 0.96 | 1.00 | 0.98 |
| Healthy (Peach) | 0.97 | 0.94 | 0.96 |
| Bacterial Spot(Peach) | 0.92 | 0.82 | 0.87 |
| Healthy(Pepper) | 0.97 | 0.94 | 0.96 |
| Bacterial Spot(Pepper) | 0.17 | 0.33 | 0.22 |
| Early Blight(Potato) | 0.94 | 1.00 | 0.97 |
| Late Blight(Potato) | 0.99 | 1.00 | 0.99 |
| Healthy(Potato) | 1.00 | 1.00 | 1.00 |
| Leaf Scorch(Strawberry) | 0.83 | 1.00 | 0.91 |
| Healthy (Strawberry) | 0.95 | 0.79 | 0.86 |
| Mosaic Virus(Tomato) | 0.96 | 1.00 | 0.98 |
| Spider Mite(Tomato) | 0.84 | 1.00 | 0.91 |
| Mold(Tomato) | 0.81 | 0.93 | 0.87 |
| Late Blight(Tomato) | 0.95 | 0.84 | 0.89 |
| Spot(Tomato) | 0.78 | 0.90 | 0.84 |
| Early Blight(Tomato) | 0.92 | 0.91 | 0.91 |
| Target Spot(Tomato) | 0.78 | 0.96 | 0.86 |
| Bacterial spot(Tomato) | 0.91 | 0.28 | 0.43 |
| Healthy(Tomato) | 0.93 | 0.93 | 0.93 |

After analyzing the results, we can see that the decompressed DCT coefficients can be fed as an input to a convolutional neural network instead of the normal RGB images for the plant leaf disease detection. Furthermore, we may deduce that CNNs that are resilient to the interrelated pixel values of the RGB picture are nearly as robust to the non-correlated decompressed DCT values.

## 5   Conclusion

We had proposed an idea of classifying plant leaf diseases based on using deep learning. This work mainly focused on feeding the DCT compressed images as input to recognize the plant leaf diseases. Working with such images has the advantage of reducing computation and data storage costs. Similarly, this approach opens a new direction for future researchers to process the data in this domain.